%% file: template.tex
\documentclass[a4paper]{article}

\usepackage{INTERSPEECH2019}
\usepackage{subfig}
\usepackage{multirow}
\usepackage{color,soul}
\usepackage{listings}
\usepackage[ruled,vlined]{algorithm2e}
\usepackage{tabularx}
\usepackage{amsfonts}
\usepackage{soul}
\usepackage[utf8]{inputenc}

\DeclareMathAlphabet{\mymathbb}{U}{BOONDOX-ds}{m}{n}

\title{ASR Model Compression Using AutoML}
\title{Automatic Speech Recognition Model Compression Using Low-Rank Matrix Factorization through Reinforcement Learning}
\title{ASR Model Compression Using SVD through Reinforcement Learning}
\title{ASR Model Compression Using SVD through AutoML}
\title{ASR Model Compression Using SVD Through Reinforcement Learning}
\title{Shrinking MoChA: Automatic ASR\\Model Compression Using Reinforcement Learning}
\title{Shrinking ASR:\\Model Compression Using Reinforcement Learning}
\title{Shrinking End-to-End ASR:\\Model Compression Using Reinforcement Learning}
\title{ShrinkML: End-to-End ASR Model\\Compression Using Reinforcement Learning}

\name{\L ukasz Dudziak$^1$*, Mohamed S. Abdelfattah$^1$*,\\Ravichander Vipperla$^1$, Stefanos Laskaridis$^1$, Nicholas D. Lane$^{1,2}$}
\address{$^1$Samsung AI Center, Cambridge\\$^2$University of Oxford\\\textit{{* Indicates equal contribution.}}}
\email{\{l.dudziak, mohamed1.a, r.vipperla, stefanos.l, nic.lane\}@samsung.com}

\definecolor{mygreen}{rgb}{0,0.6,0}
\definecolor{mygray}{rgb}{0.5,0.5,0.5}
\definecolor{mymauve}{rgb}{0.58,0,0.82}

\newcommand{\til}{{\fontfamily{ptm}\selectfont\texttildelow}}
\newcommand{\xx}{$\times${ }}

\newcommand{\figvs}[4]{\begin{figure}[!t]
\centering
\includegraphics[width=#1\columnwidth,keepaspectratio,#3]{#2}
\caption{#4}
\label{#2}
\end{figure}}

\newcommand{\comment}[1]{}

\newcommand{\hlc}[2][yellow]{{\sethlcolor{#1} \hl{#2}}}
\newcommand{\addcomment}[2]{\hlc[green]{#1} \hl{#2}}

\comment{
\addtolength{\floatsep}{-5pt}
\addtolength{\dblfloatsep}{-20pt}
\addtolength{\abovecaptionskip }{-4pt}
\addtolength{\parskip }{-0.5pt}

}

\begin{document}
\maketitle
\begin{abstract}
End-to-end automatic speech recognition (ASR) models are increasingly large and complex to achieve the best possible accuracy.
In this paper, we build an AutoML system that uses reinforcement learning (RL) to optimize the per-layer compression ratios when applied to a state-of-the-art attention based end-to-end ASR model composed of several LSTM layers.
We use singular value decomposition (SVD) low-rank matrix factorization as the compression method.
For our RL-based AutoML system, we focus on practical considerations such as the choice of the reward/punishment functions, the formation of an effective search space, and the creation of a representative but small data set for quick evaluation between search steps.
Finally, we present accuracy results on LibriSpeech of the model compressed by our AutoML system, and we compare it to manually-compressed models.
Our results show that in the absence of retraining our RL-based search is an effective and practical method to compress a production-grade ASR system. 
When retraining is possible, we show that our AutoML system can select better highly-compressed seed models compared to manually hand-crafted rank selection, thus allowing for more compression than previously possible.
\end{abstract}
%


\section{Introduction}
\input{intro}


\section{ASR Model Compression}

\input{asr_model}

\section{AutoML for SVD Rank Selection}

\input{automl}


\section{Results}

\input{results}

\section{Conclusion}
\input{conclusion}

\section{Acknowledgements}
\input{acknowledgements}

\bibliographystyle{IEEEtran}
\bibliography{mybib}


\end{document}

%% file: intro.tex
\comment{
\begin{itemize}
    \item Motivate the need to have a compression methodology for ASR models to deploy on constrained systems such as mobile. 
    \item This paper should be a practical walk-through of our model compression methodology with notes on how to do AutoML well for SVD-based approximation.
    \item We show how automl can be useful in both compressing a model without retraining, or selecting the seed model for retraining with aggressive compression.
\end{itemize}
}

End-to-end automatic speech recognition (ASR) models have outperformed traditional ASR systems. 
In particular, models with attention mechanism~\cite{Bahdanau2015NeuralMT} have produced state-of-the-art results in speech recognition and translation application domains~\cite{Chan2016ListenAA,Chorowski2014EndtoendCS,Luong2015EffectiveAT}. 
These models improve upon traditional systems, partially because they include acoustic models, language models and pronunciation dictionary jointly learned together. 
Furthermore, models with character/subword output units are especially useful in low latency deployments such as  mobile phones and can produce good results with a beam-search decoding mechanism~\cite{googleOnDeviceStreamingASR}.

One challenge when encompassing all components of ASR into a single model is the need for larger model capacity and thereby deeper neural networks with several recurrent layers. 
This makes it harder to achieve real-time inference, especially with on-device deployments.
One approach to reduce model size is to apply model compression techniques~\cite{pang_compression_2018}.
This is often done manually and may include many trial-and-error experiments until a reasonable compression scheme is found.
However, more recently, automated search-based techniques have been employed to optimize the compression of deep neural networks~\cite{han_automl_compress}.
More specifically, \textit{AutoML} has been used to select the per-layer compression ratios to optimize a model globally.

\comment{
One obvious approach to compress models with several large matrices is to use the SVD based low-rank factorization technique which was shown to be practically very useful~\cite{someone}. 
The ranks of LRF are typically chosen manually or by using a crude search over ranks for a given matrix. This approach becomes unfeasible for deep networks especially with recurrent layers. 
It is also not clear if optimizing one matrix at a time would lead to an overall optimal system in terms of WER. 
There seems to little prior work in addressing this issue. 
}

In this paper, we use reinforcement learning to select the per-layer compression ratios based on matrix approximation using singular value decomposition (SVD)~\cite{sourav}.
We base our work on a recently proposed end-to-end ASR model with attention mechanism \cite{Zeyer2018ACA, Zeyer2018ImprovedTO} that has obtained state-of-the-art results on LibriSpeech. 
We focus on practical aspects when designing the AutoML compression system in the specific context of ASR models.
We show experimentally that our approach is superior to manual model compression for both modest and aggressive compression targets, in the absence or presence of retraining.
In the following sections, we outline the ASR model and present insights into compressibility of layers in such encoder-attention-decoder models, describe the AutoML framework and present a data subsampling technique to speed up the search process, and finally delve into experimental results that prove the efficacy of our system, corroborated by on-device measurements on a mobile phone.

%% file: asr_model.tex
\figvs{0.77}{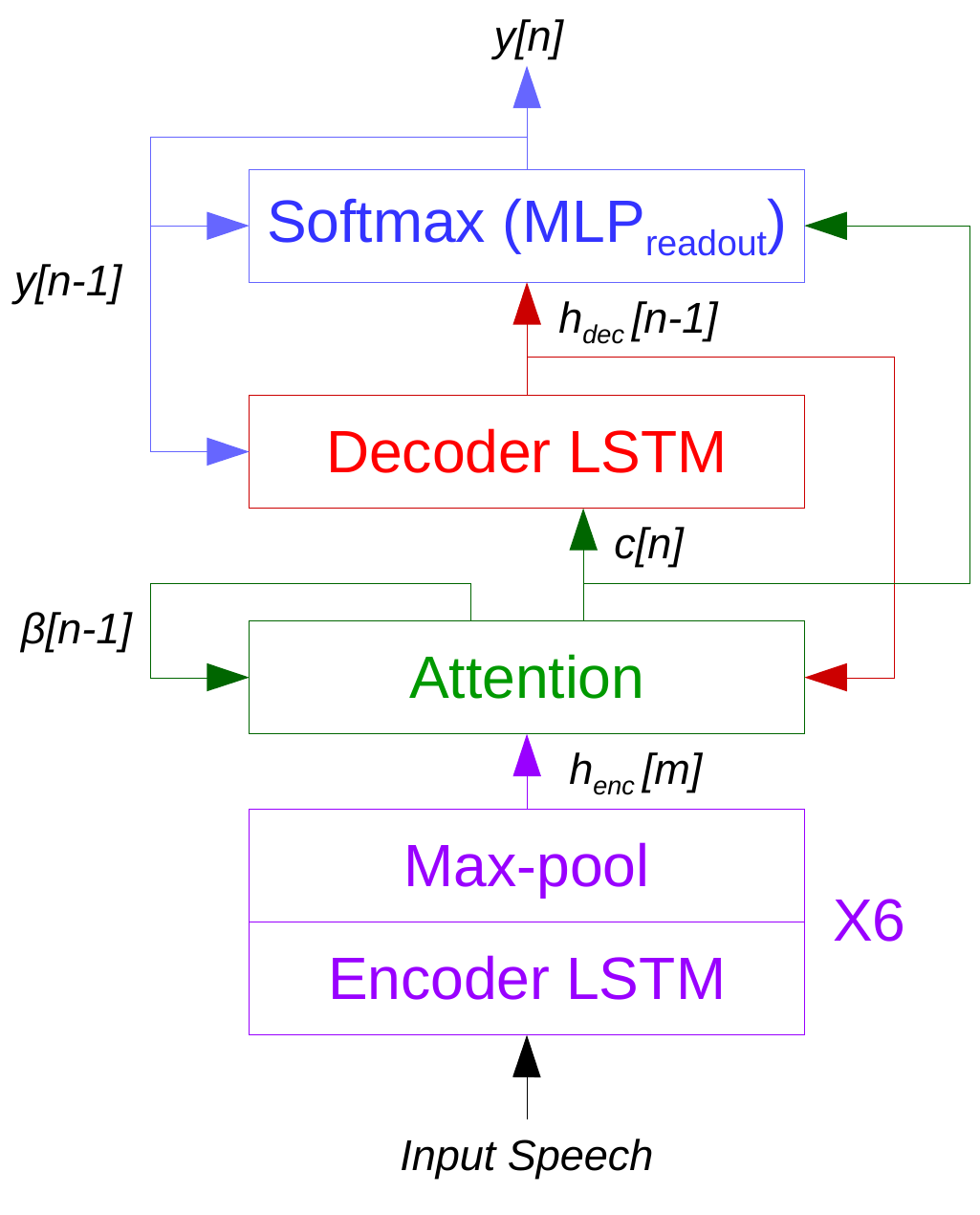}{trim = 0 20 0 20}{End-to-end ASR model architecture.}
%

%
%

\subsection{ASR Model}
\label{sec_model}

We use an end-to-end attention based ASR model~\cite{asru1,asru2} with an architecture similar to the one proposed in \cite{Zeyer2018ImprovedTO} as depicted in Fig.~\ref{model}.
The main differences in our model are the use of 1) unidirectional LSTMs in place of bidirectional LSTMs and 2) monotonic chunk-wise attention~\cite{chiu2018monotonic}. 
These changes were made to allow the model to decode speech in streaming mode, which is required for mobile applications. 
As a result, the model also has a higher baseline word error rate (WER) as compared to the results published in \cite{Zeyer2018ImprovedTO}.
The unit size for all LSTM layers in the network is 1024 and the output layer comprises 10000 sized byte-pair encoded sub-word units~\cite{Sennrich2016bpe, isck1}. 
The model was trained on LibriSpeech training set~\cite{Panayotov2015Libri} with MFCC front-end using Tensorflow \cite{tensorflow2015-whitepaper} and the RETURNN framework~\cite{zeyer2018returnnAcl}. 
A time reduction factor of 8 was used via the maxpool layers interleaved between the first four encoder LSTM layers. 
In this work we have not used a separate language model for further reduction of WER.

\subsection{Low-Rank Matrix Approximation}

\comment{
\begin{itemize}
    \item Introduce SVD approximation
    \item Show potential for model compression through SVD applied on the matrices.
    \item Display some of the energy curves (if we can draw conclusions from them).
\end{itemize}
}

Our primary target is to build a generic reinforcement learning (RL) based AutoML system that automatically optimizes the per-layer compression ratios for ASR models.
We decide to choose low-rank matrix factorization, and specifically, singular-value decomposition (SVD) to compress the weight matrices, since it has proven to be an effective method for compressing similar models~\cite{xue_restructuring_nodate,xue14,ming17,tucker_model_2016,prabhavalkar_compression_2016,povey_semi-orthogonal_2018}.
However, we design our system such that it is generalizable to different compression methodologies.
We refer the reader to prior work for further reading on SVD, but a summary is given below.
We factorize a matrix $M$ into three matrices $U$, $\Sigma$ and $V$:

\vspace{-0.2cm}
\begin{equation}
    M = U\Sigma V^T
\end{equation}

\noindent where $\Sigma$ is a diagonal matrix which consists of the singular values (typically in descending order of magnitude) that uniquely identify $M$.
We compress $M$ by removing the lower-magnitude values in $\Sigma$ -- we refer to the number of remaining element as a \textit{factorization rank} and denote it with $k$.
We select $k$ in such a way that it preserves a given \textit{energy}, where energy refers to the normalized summation of the remaining singular values.
We then remove corresponding columns from matrices $U$ and $V$, creating a set of three new matrices $U'$, $\Sigma'$ and $V'$ which approximate original matrix $M$.
Finally, we combine $\Sigma'$ with $V'$ so that we can replace $M$ according to the following equation:

\vspace{-0.2cm}
\begin{equation}
    M \approx U'V^* = U'\Sigma' {V'}^T
\end{equation}

\noindent If $M$ is an $m\times n$ matrix, then $U'$ and $V^*$ have dimensions $m\times k$ and $k\times n$ respectively, and we get a theoretical speedup and model size reduction:

\vspace{-0.3cm}
\begin{equation}
\label{eqn_speedup}
    Speedup = \frac{m\times n}{k\times (m+n)}
\end{equation}

\comment{TODO: mention that usually, when compressing more than one layer, $k$ values are selected manually, basic on empirical results -- this is the motivation for us to use AutoML. The same problem appears not only in SVD compression but also in different compression/architectural techniques which are based on the bottleneck structure, so our automl (or just some tricks we did to make it faster) should be useful for different approaches as well. Make sure it doesn't repeat whatever was written at the beginning of 2.2, maybe restructure the text a little bit.}

%% file: automl.tex
\figvs{1}{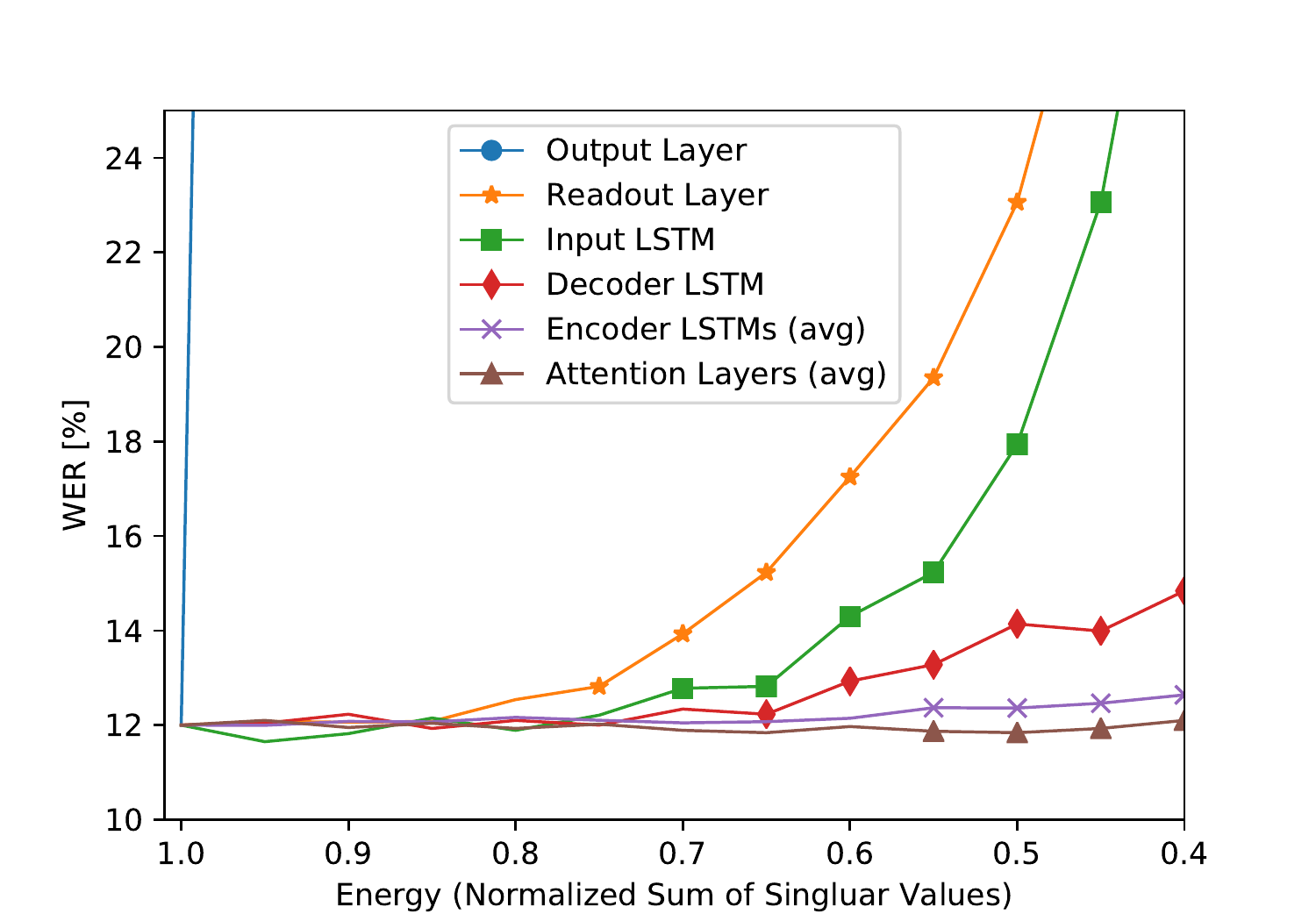}{trim = 0 0 0 30}{Single-layer compression gives an indication of the relative sensitivity of the model to each layer, and instructs the acceptable compression ranges for our RL-based search.}

Fig. \ref{single_layer} plots the WER degradation when each of the layers are factorized separately -- as shown, the model accuracy is much more sensitive to changes in the input and output layers, but less so to intermediate encoder LSTM cells, attention and decoder layers.
Selecting the ranks per layer for the 18 matrices in the model is a hard problem because of the large number of combinations.
For instance, if we discretize compression range for each layer to 5 rank options, we have $5^{18}=3.8\times10^{12}$ possible factorizations.
We use reinforcement learning to navigate that search space and find the best compression schemes.

\subsection{AutoML System}


We use an RL-based system similar to the one proposed by \cite{zoph_neural_2016} after  slightly adapting it for the model compression task. 
More specifically, in our case an agent is responsible for deciding the factorization scheme which would: \textit{a)} guarantee a predefined speedup, and \textit{b)} minimize WER of the factorized model.
The scheme in this context is a series of decisions, with each decision representing a compression ratio for an individual layer.
Both the list of layers and the set of discrete compression levels per layer are provided a priori to the search.
We call the set of all possible factorization schemes a \textit{search space} and the set of all parameters which modify it \textit{hyperparameters of the search space}.
More formally, our search space can be defined as a 2-dimensional matrix: $S \in \mathbb{Z}^{l\times d}$, where $S_{i,j}$ represents a rank $k$ to use when factorizing the $i$-th layer according to the $j$-th option.
From this search space, the agent selects a factorization scheme $s \in \mathbb{Z}^{l}$ by choosing one value from each row of matrix $S$.
The selection is done by sampling $l$ probability distributions over $d$ options which are produced by a trainable policy $\pi(\theta)$.

In our system, this policy is modeled with a single LSTM layer with 100 input units and 100 hidden units which takes a sequence of $l$ inputs (one for each layer) -- each element of output sequence is then passed to its individual fully-connected layer with $d$ units, to match its length with the number of available decisions, followed by softmax.
The final output is $D \in \mathbb{R}^{l\times d}$ where each vector $D_{i} \in \mathbb{R}^{d}$ represents probability distribution of selecting different factorization options for the $i$-th layer.
We reward the agent depending on the WER of the model when compressed according to the proposed scheme, and use policy gradient to update $\theta$. 

\comment{
\begin{algorithm}
\LinesNumbered
\SetAlgoLined
\KwIn{Policy weights $\theta$, set of explored points $\mathbb{W}$, search space $S$}
\KwOut{Updated $\theta$ and $\mathbb{W}$}
$D \leftarrow \pi(\theta)$ \\
$\hspace{0.1cm} s \leftarrow \mymathbb{0}_{l}$ \\
$\hspace{0.1cm} p \leftarrow \mymathbb{0}_{l}$ \\
\For{$i$ $\leftarrow$ $0$ \KwTo $l$}{
    $j \sim D_{i}$ \\
    $s_{i}$ $\leftarrow$ $S_{i,j}$ \\
    $p_{i}$ $\leftarrow$ $D_{i, j}$
}
$w$ $\leftarrow$ compress model according to $s$ and evaluate WER \\
$\theta$ $\leftarrow$ update $\theta$ using $\nabla(-\log(\prod_{i=0}^{l}p_{i})\mathcal{R}(w))$ \\
$\mathbb{W}$ $\leftarrow$ $\mathbb{W} \cup \{ (s, w) \}$
\caption{A single step of our RL-based search.}
\label{alg:search}
\end{algorithm}
}

\subsection{AutoML System Tuning for Compression}

\begin{algorithm}[t]
\LinesNumbered
\SetAlgoLined
\KwIn{Policy weights $\theta$, set of explored points $\mathbb{W}$, search space $S$, target speedup $a_{t}$}
\KwOut{Updated $\theta$ and $\mathbb{W}$}
$D \leftarrow \pi(\theta)$ \\
$\hspace{0.1cm} s \leftarrow \mymathbb{0}_{l}$ \\
$\hspace{0.1cm} p \leftarrow \mymathbb{0}_{l}$ \\
\For{$i$ $\leftarrow$ $0$ \KwTo $l$}{
    $j \sim D_{i}$ \\
    $s_{i}$ $\leftarrow$ $S_{i,j}$ \\
    $p_{i}$ $\leftarrow$ $D_{i, j}$
}
$a$ $\leftarrow$ estimate speedup achievable by $s$ \\
\eIf{$a < a_t$}{
    $\Delta_a \leftarrow a_t - a$ \\
    $r \leftarrow \mathcal{R}_v(\Delta_a)$ \\
}{
    $w$ $\leftarrow$ compress model according to $s$ and evaluate WER on a proxy dataset \\
    $r$ $\leftarrow$ $\mathcal{R}(w)$ \\
    $\mathbb{W}$ $\leftarrow$ $\mathbb{W} \cup \{ (s, w) \}$ \\
}
$\theta$ $\leftarrow$ update $\theta$ using $\nabla(-\log(\prod_{i=0}^{l}p_{i})r)$ \\

\caption{A single step of the proposed search.}
\label{alg:search_fast}
\end{algorithm}

One of the main challenges for many AutoML systems is reducing the time required to produce useful results. 
RL-based search is slow because evaluation of a proposed action (in our case, running evaluation on a validation set) takes significantly more time than proposing a new action.
In this work we address this problem and propose a number of techniques which we have successfully used to significantly reduce evaluation time. This way, we speed up the search, while still providing representative feedback to the RL agent.



To quickly reject unprofitable points, we estimate speedup of a proposed scheme ($a$) and compare it to the predefined target ($a_t$), as shown in Algorithm \ref{alg:search_fast}. 
If it falls below the threshold, we \textit{punish} the agent using a dedicated reward function $\mathcal{R}_v$, otherwise the scheme is accepted for full evaluation and we reward the controller according to returned WER using reward function $\mathcal{R}$. 
We observed that combined usage of the two reward functions makes the search converge to a certain area of the speedup-WER plane -- as presented in Fig.  \ref{search_points} -- with $\mathcal{R}$ pushing points towards smaller WER and $\mathcal{R}_v$ pushing above $a_t$. 
However, both functions need to be tuned relative to each other, as well as independently, to obtain good results. 
In our work we empirically found that when targeting conservative $a_t$ values relative differences in WER between the original model ($w_b$) and its compressed versions ($w$) might be too small to make the agent discriminate between good models with similar WERs. 
Therefore, we decided to use exponential function to add additional emphasis to the difference:

\vspace{-0.2cm}
\begin{equation}
    \label{eq:reward1}
    \mathcal{R}(w) = -\exp(w-w_{b})
\end{equation}

\noindent However, the range of this function is too big when targeting high speedups $a_t$, because the WER range is much larger, so we used a slightly modified version for such aggressive searches:

\vspace{-0.2cm}
\begin{equation}
    \label{eq:reward2}
    \mathcal{R}(w) = -\exp(\sqrt{\frac{w}{w_{b}}})
\end{equation}

\noindent For both aggressive and conservative compression we observed that a simple linear function works well for $\mathcal{R}_v$, with parameters $-100$ and $-10$ tuned to make its values visibly worse than $\mathcal{R}$, even when $\Delta_a \rightarrow 0$.

\vspace{-0.2cm}
\begin{equation}
    \label{eq:punishment}
    \mathcal{R}_{v}(\Delta_a) = -100\Delta_a-10
\end{equation}


\comment{
\begin{itemize}
    \item Explain early exit (based on speedup).
    \item Talk about the design of the reward/punish functions.
    \item How do we make the search fast and practical? It has to be faster than retraining to be useful
    \item Guided search-space selection based on empirical evaluation of the space.
\end{itemize}
}

\comment{
\subsubsection{Search-Space Reduction}

The first one, although quite obvious, must be approached with care since limiting the search space to make the search faster also limits the agent in available options, thus possibly prohibiting it from finding better results. \addcomment{We used information about WER degradation when compressing individual layers independently (see Figure X) to identify reasonable set of layers (hyper parameter $l$) and energy ranges for each of them (hyperparameter $d$). Specifically, we excluded output and readout layers and either exclude or set conservative compression range for input LSTM (depending on overall aggressiveness) -- other layers were treated similarly, with attention layers having slightly more aggressive ranges.}{reduce with 3?}
}

\subsection{Small Dataset Creation}
\comment{
\begin{itemize}
    \item Explain selection of small dataset.
    \item Show how correlation changes on random/targeted small datasets of the same size. 
    \item Present some findings from the search on targeted dataset?
    \item Hopefully show how the search on a small set is just as good as a large one but 5X as fast.
\end{itemize}
}

\figvs{1}{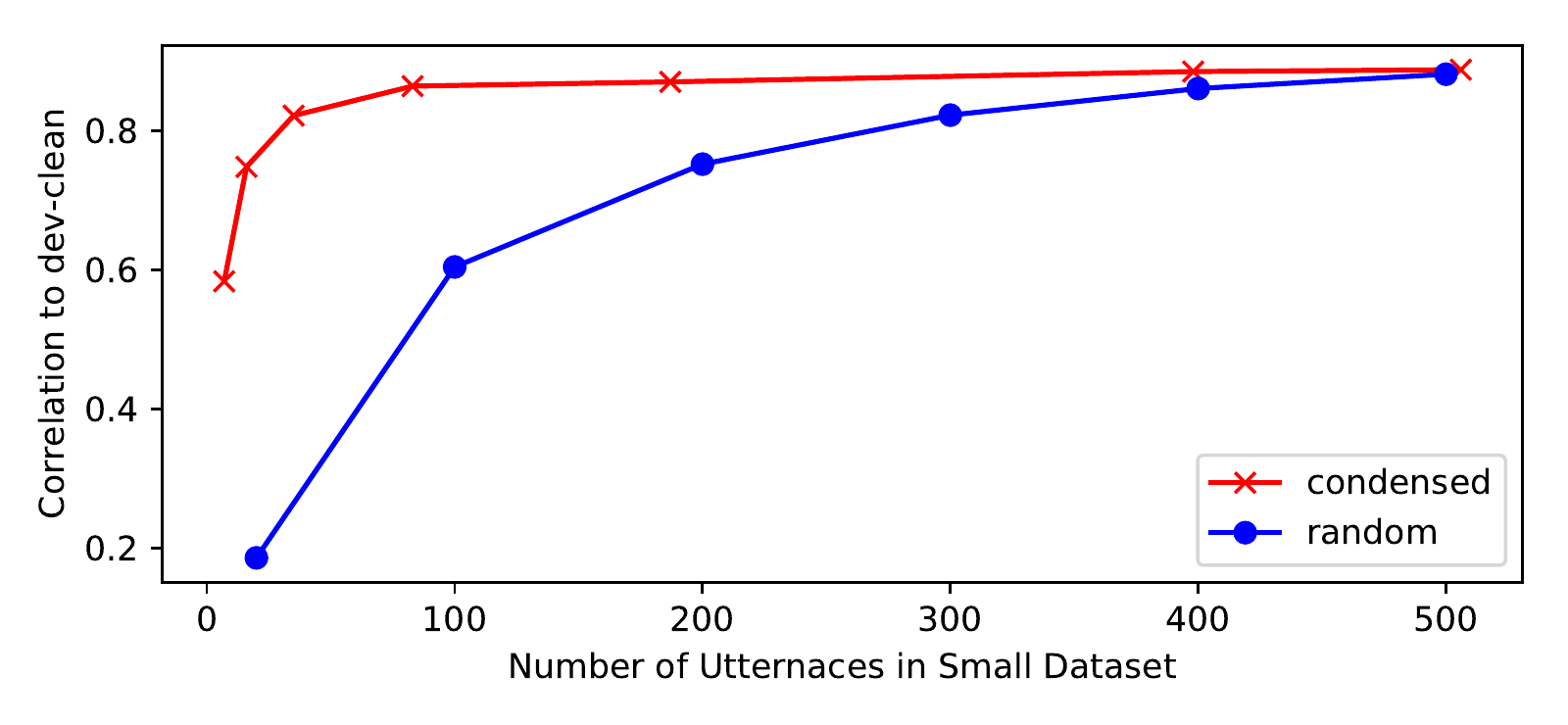}{trim = 0 20 0 5}{Correlation of small datasets with the full-size validation sets. Our ``condense" algorithm is more successful in creating smaller representative datasets for use with AutoML.}

\comment{
We start by using the validation sets from LibriSpeech to evaluate WER for models proposed by AutoML, however, we quickly realize that this is too slow and we investigate if we can find a smaller set that can represent the validation sets but run much faster}

LibriSpeech validation sets consist of 5567 utterances split among \textit{dev-clean} and \textit{dev-other}. 
Using the validation sets to evaluate WER for each model proposed by AutoML was too slow and led us to investigate if a representative subset could be found.
First, we tried randomly selecting utterances from the validation sets.
However, we observed experimentally that this did not work well with small random sets -- if we minimize WER on the random set, this does not always translate into a lower WER on the full-size validation or test sets.
Instead, we used our ``condense" algorithm in Listing~\ref{list2} to select the utterances that were most representative of the validation sets.

\begin{small}
\begin{lstlisting}[language=python, caption=Heuristic to find ``condensed" datasets in Fig.~\ref{dataset_correlation}., label=list2]
#wer_avg_all is a list containing WER on whole dev set w.r.t. cohort models
for utterance in validation_sets:
    wer_utt= []
    for c_model in cohort_models:
        wer_utt.append(
            compute_wer(utterance,c_model))
    correl[utterance] = 
        correlation(wer_utt, wer_avg_all)

#filter utterances < min_correl and min_length
for utterance in validation_sets:
    if correl[utterance] > correl_min:
        new_set.append(sample)
\end{lstlisting}
\end{small}

As Listing~\ref{list2} shows, we find the WER per utterance from different cohort models.
These models are simple variations of our baseline model presented in Section~\ref{sec_model} -- we use 8 cohort models trained with different number of layers, layer sizes in the encoder, and few variations of SVD-based compression schemes.
We compile the WERs of each utterance when decoded with each of these models, and correlate that to the WERs of the entire validation set with the same models.
We then choose the utterances that correlate highly in WER with the entire set across all the cohort models.
As Fig.~\ref{dataset_correlation} shows, our ``condensed" datasets correlate much better than ``random" sets, especially with very small dataset sizes.
With ${correl}\_{min}=0.95$ we created an 83-sample dataset for use with AutoML -- this is 67\xx smaller than the validation sets but was highly-correlated with the full set as Fig.~\ref{dataset_correlation} shows.
We used it as a drop-in replacement for AutoML thus making the whole system $>$10\xx faster overall.

%% file: results.tex
\begin{figure*}[t]
\centering
\subfloat[Steps 0 to 300.]{
   \includegraphics[width=0.66\columnwidth,keepaspectratio, trim=45 0 10 45]{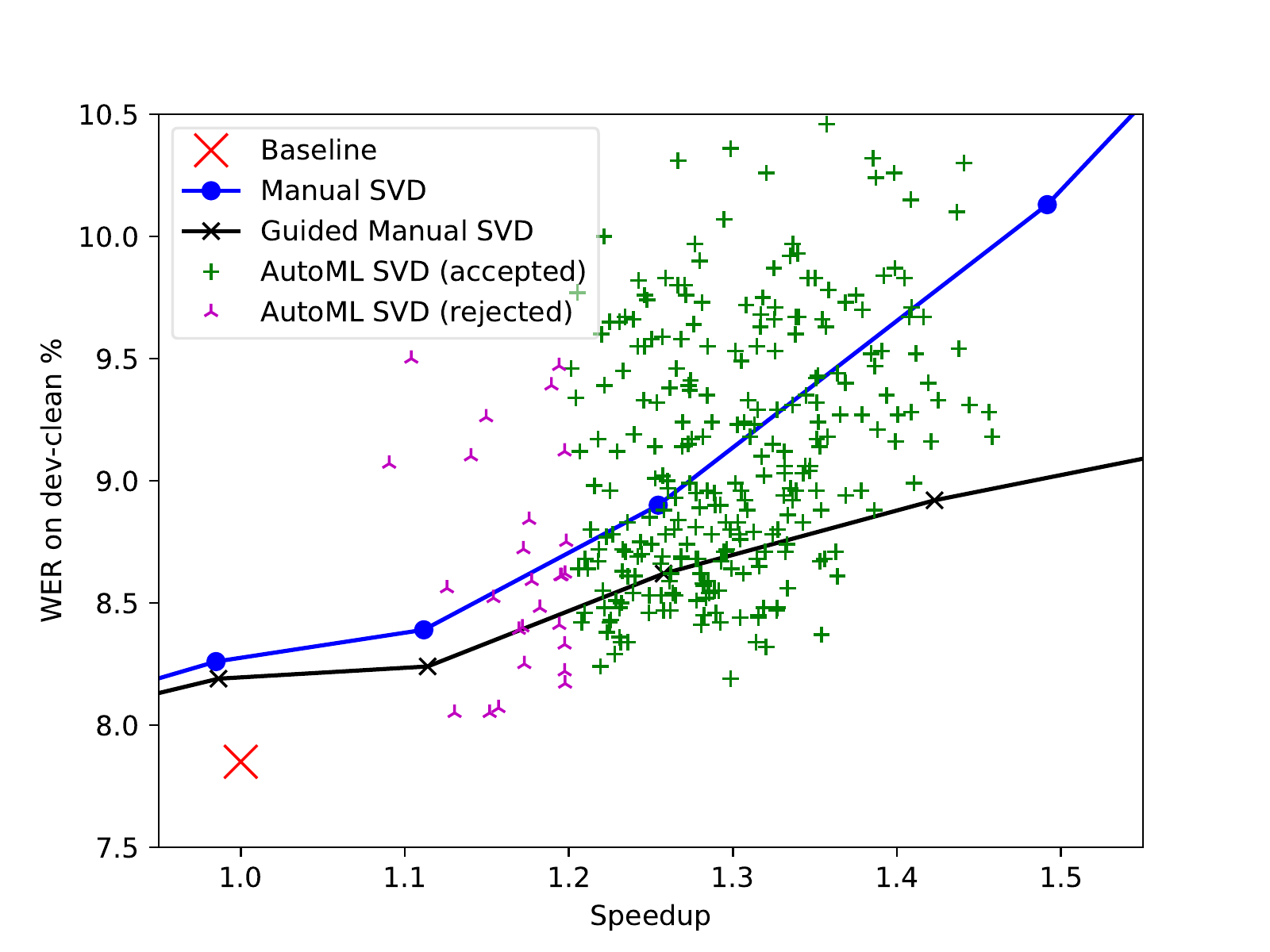}
   \label{credits_3m}
 }
\subfloat[Steps 600 to 900.]{
   \includegraphics[width=0.66\columnwidth,keepaspectratio, trim=45 0 10 45]{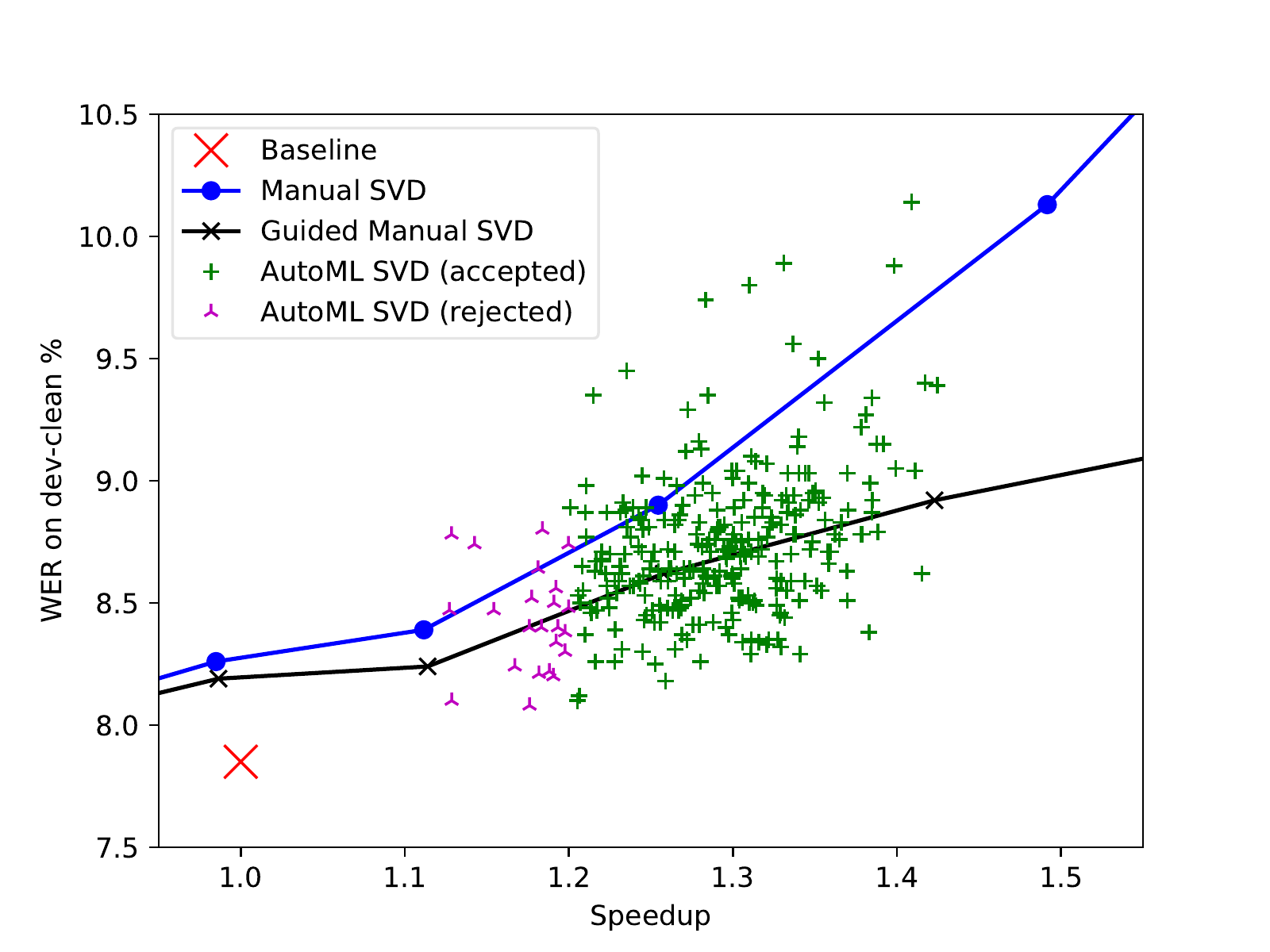}
   \label{credits_6m}
 }
\subfloat[Steps 1600 to 1900.]{
   \includegraphics[width=0.66\columnwidth,keepaspectratio, trim=45 0 10 45]{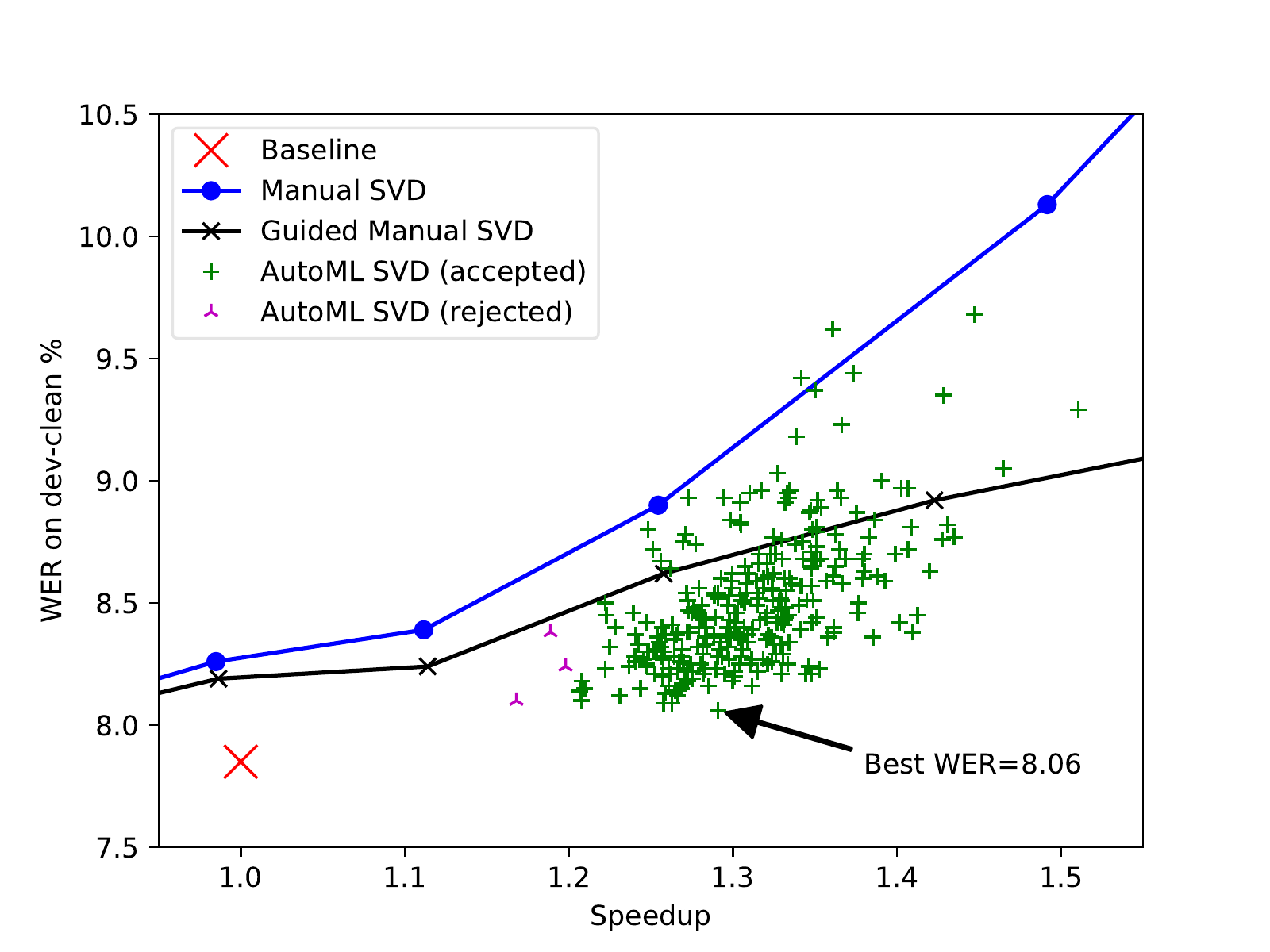}
   \label{credits_9m}
 }
\caption{Comparison of the RL search with na\"ive manual SVD compression (all layers with equal \textit{energy}), and ``guided" manual SVD (excludes sensitive layers identified in Fig.~\ref{single_layer}). As the search progresses, results that outperform manual compression are explored.}
\label{search_points}
\end{figure*}

We use AutoML for two experiments.
First, we want to find a modestly-compressed model with minimal WER loss in the absence of retraining.
We believe that full training data is not always available, especially in systems where all of the training data is not in one place~\cite{federated_learning}, or when app developers are using pretrained models and do not have access to training sets.
Second, we use AutoML to find the best aggressively-compressed seed model for retraining.
In this case we maximize compression, and rely on retraining to recover the accuracy.

\subsection{Compression without Retraining}

We launched two searches, \textit{slow} and \textit{fast}, both targeting speedup $a_t = 1.2$, and using Equations~\ref{eq:reward1} and~\ref{eq:punishment}.
The slow search evaluated models on the entire \textit{dev-clean} dataset and selected compression ratios for each layer from the same range.
The fast search used the condensed dataset described in the previous section and had per-layer compression ranges adjusted according to the information from Fig.~\ref{single_layer}.
To extract the best model, for each search independently we gathered all explored configurations and selected top-5 (as evaluated by the search) which were then evaluated on the ``test" datasets from LibriSpeech (\textit{test-clean} and \textit{test-other}).
The best model reported here is the best of the 5 on \textit{test-clean}.
As shown in Table~\ref{tab:no_retrain}, the fast search was able to find an equally-good compression scheme much faster.
Both searches were also able to find better compression schemes than the hand-crafted manual ones as highlighted in Fig~\ref{search_points}.

\begin{small}
\begin{table}[th]
    \centering
    \caption{Evaluation of the best models found by both searches. \textit{Steps} mean at what step during the search a model was found.}
    \begin{tabular}{|l|c|c|c|c|}
         \hline
                        & \multicolumn{2}{c|}{AutoML}   & Manual    & \multirow{2}{*}{Baseline} \\\cline{2-3}
                        & Slow      & Fast              & Guided    &                           \\\hline
         test-clean     & 8.35      & 8.34              & 8.67      & 8.29                      \\\hline
         test-other     & 21.61     & 21.33             & 22.04     & 21.13                     \\\hline
         step           & 1626      & 2363              &  \multicolumn{2}{c|}{\multirow{2}{*}{NA}} \\\cline{1-3}
         GPU hours      & 635       & 74                &  \multicolumn{2}{c|}{} \\\hline
    \end{tabular}
    \label{tab:no_retrain}
\end{table}
\end{small}

\vspace{-0.5cm}

\subsection{Aggressive Compression With Retraining}
\label{sec_retrain}

\comment{
\begin{itemize}
    \item Talk about how retraining is a very effective method of recovering accuracy as previous work has shown.
    \item Does it help if we find the highly-compressed model from AutoML compared to manual? The idea is that we achieve the same compression ratio but with a better configuration. Maybe it reduces the number of epochs we need to retrain for?
\end{itemize}
}

Previous work has repeatedly proven that SVD-based compression is very effective in attaining speedup while maintaining model accuracy when retraining is available~\cite{xue_restructuring_nodate,ming17,prabhavalkar_compression_2016}.
We use our ``guided manual" method of compression to test the limits of compression with retraining. 
We train for 50 \textit{epochs} over LibriSpeech with an epoch split of 20 -- this means we go through the training data 2.5 times.
As Fig.~\ref{retraining} shows, we achieved up to \til3\xx compression without degrading WER on test-clean. 
Can we push this any further using our AutoML system?
To answer this question, we set our speedup threshold to 3.7\xx and launched our \textit{fast} AutoML search for only a few hours, we then used the best-found model as the seed model for retraining and the results are shown in Fig.~\ref{retraining}.
The model found by AutoML achieved 3.75\xx speedup and had approximately half the WER before retraining when compared to the manually-compressed model at the same speedup.
After retraining, we were able to recover almost full accuracy, as shown in Fig.~\ref{retraining}, therefore, we believe this is an effective systematic method to selecting highly-compressed models for retraining.
After 8-bit quantization, the model size is compressed to 24~MB from the 32-bit compressed model size of 72~MB.

\figvs{1}{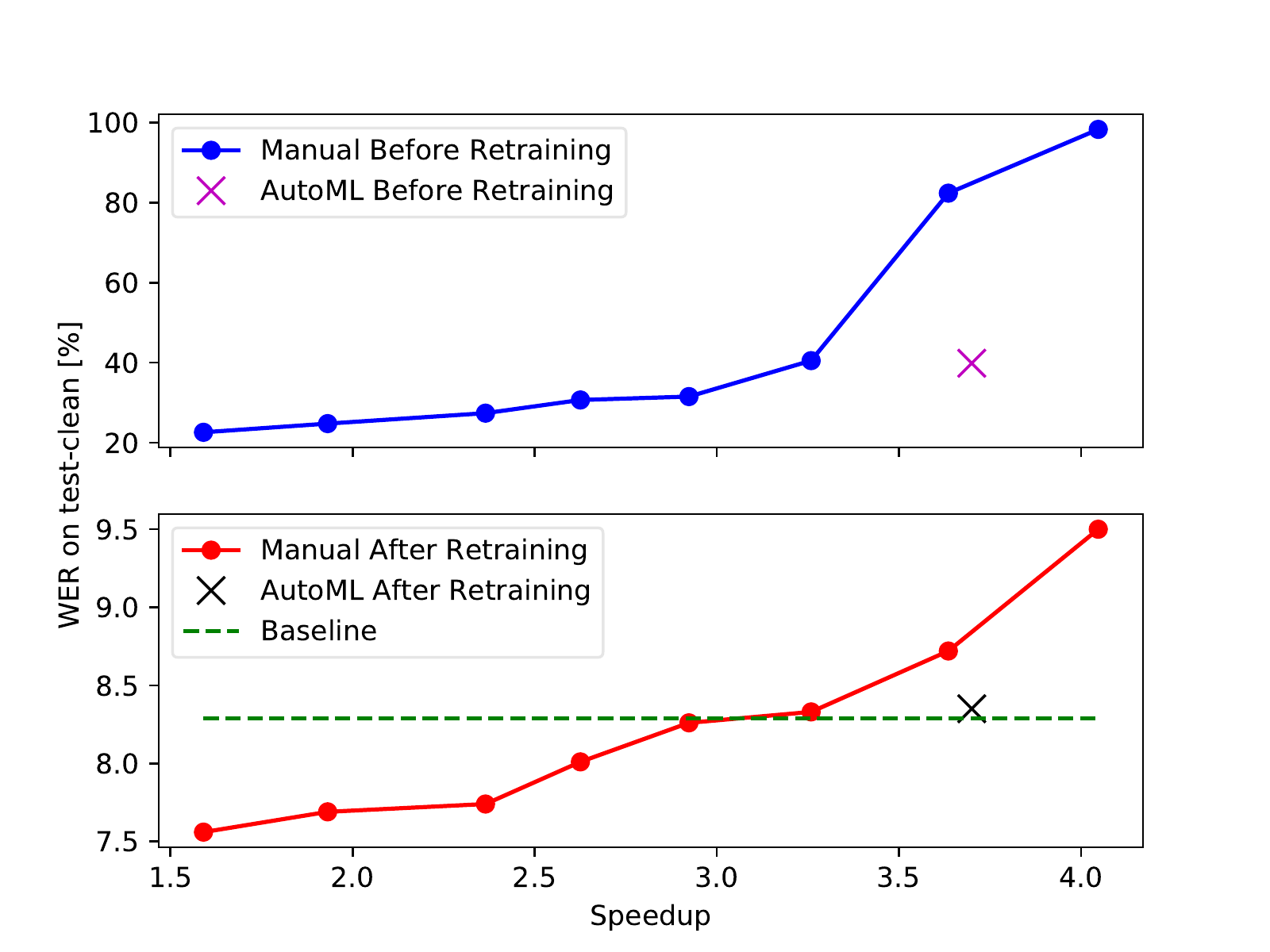}{trim = 0 7 0 35, clip}{WER before/after retraining for ``guided manual" and AutoML compressed models.}

\comment{
\begin{table}[th]
  \caption{Word Error Rate [\%] for aggressively compressed models (with speedup \texttildelow3$\times$) when seed model is created manually or found by AutoML.}
  \label{agg_table}
  \centering
  \begin{tabular}{|c|l|c|c|c|c|c|}
    \hline
    \multicolumn{2}{|c}{\multirow{2}{*}{Model}}  & \multicolumn{2}{|c|}{dev} & \multicolumn{2}{|c|}{test} \\\cline{3-6}
                \multicolumn{2}{|c|}{}                    & clean & other & clean & other  \\\hline
    Before                                   & Manual     &       &       &       &        \\\cline{2-6}
    Retraining                               & AutoML     &       &       &       &        \\\hline
    After                                    & Manual     &       &       &       &        \\\cline{2-6}
    Retraining                               & AutoML     &       &       &       &        \\\hline
    \multicolumn{2}{|c|}{Baseline}                        & 7.85  & 20.06 & 8.29  & 21.13  \\\hline
  \end{tabular}
\end{table}
}

\subsection{On-Device Measurements and Considerations}

%
\figvs{1}{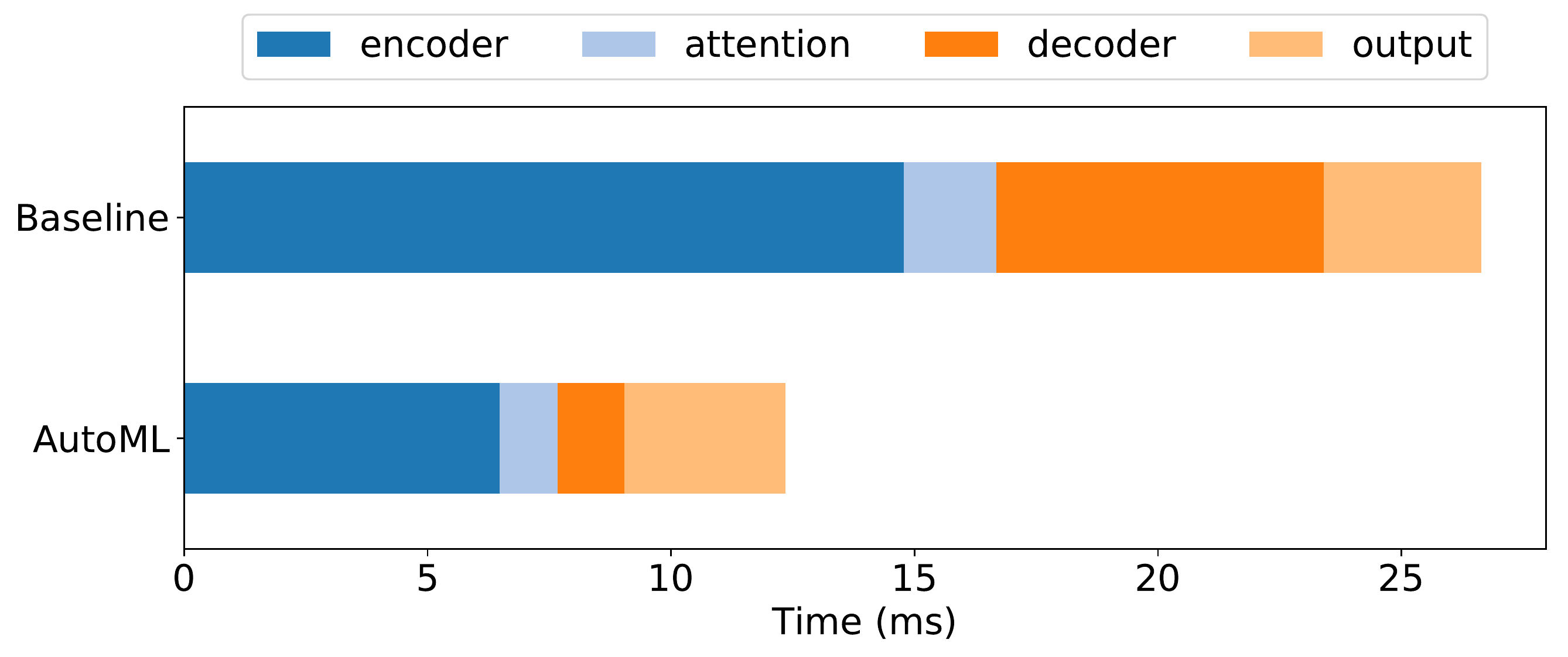}{}{Runtime on Qualcomm Snapdragon 845 chipset.}

So far, we have been using Equation~\ref{eqn_speedup} as a proxy for speedup, however, we measured the actual on-device runtime of our retrained model from Section~\ref{sec_retrain} to validate our estimate. 
We evaluate on a Qualcomm Snapdragon 845 development board, using TFLite~\cite{tflite} on CPU.
As depicted in Fig.~\ref{StackedBarSVDTimesCoarse}, we acquire a significant speedup of 2.16\xx with our SVD-based compressed model over the baseline -- especially in the encoder and decoder parts of the network. 
However, this is much lower than the estimated 3.74\xx shown in Fig~\ref{retraining}.
We shall investigate this discrepancy in future work but mention this here as a note that theoretical and measured speedups often vary greatly.


%% file: conclusion.tex
In this work, we presented an AutoML framework to push the boundaries of SVD-based ASR model compression beyond what is possible manually.
We improved upon the WER attainable by manual compression when retraining is not possible.
Even when we could retrain, we have shown that AutoML can improve the compression ratio, and therefore speedup, of  our ASR model without any loss of accuracy.
In the future we aim to use AutoML to optimize and mix different compression techniques, and we hope to make the estimated speedup/accuracy in such a system more faithful to the actual results. 

%% file: acknowledgements.tex
MoChA within the experimental framework was implemented by Dr. Chanwoo Kim and his team in Samsung research. We would like to thank them for their support and feedback.